\documentclass[letterpaper, 10 pt, journal, twoside]{IEEEtran}

\ifCLASSINFOpdf
\usepackage[pdftex]{graphicx}
\else
\fi

\usepackage{amsmath}
\usepackage{amssymb}
\usepackage{times}
\usepackage{algorithmic}
\usepackage{algorithm}
\usepackage{url}
\usepackage{multirow}
\usepackage[dvipsnames]{xcolor}
\usepackage{eso-pic, rotating, graphicx}

\newcommand{\first}[1]{{\color{blue} \textbf{#1}}}
\newcommand{\second}[1]{{\color{OliveGreen} #1}}
\newcommand{\third}[1]{{\color{red} #1}}

\def\mat#1{\mathchoice{\mbox{\boldmath $\displaystyle\tt#1$}}
{\mbox{\boldmath$\textstyle\tt#1$}}
{\mbox{\boldmath$\scriptstyle\tt#1$}}
{\mbox{\boldmath$\scriptscriptstyle\tt#1$}}}

\def\vect#1{\mathchoice{\mbox{\boldmath $\displaystyle\bf#1$}}
{\mbox{\boldmath  $\textstyle\bf#1$}}
{\mbox{\boldmath  $\scriptstyle\bf#1$}}
{\mbox{\boldmath  $\scriptscriptstyle\bf#1$}}}

\def\v0{{\vect 0}}

\def\va{{\vect a}}

\def\vf{{\vect f}}

\def\vh{{\vect h}}

\def\vn{{\vect n}}
\def\vp{{\vect p}}

\def\vu{{\vect u}}

\def\vx{{\vect x}}
\def\vy{{\vect y}}
\def\vz{{\vect z}}

\def\vD{{\vect D}}

\def\v0{{\vect 0}}

\def\mDelta{{\mat\mDelta}}

\def\mB{{\mat B}}

\def\m1{{\mat 1}}

\def\cF{\mathcal{F}}

\def\cF{\mathcal{F}}

\def\cL{\mathcal{L}}

\def\cP{\mathcal{P}}

\def\cS{\mathcal{S}}

\def\cT{\mathcal{T}}

\AddToShipoutPicture{\put(600,740){\rotatebox{-90}{\scalebox{1}{\color{gray}\copyright 2021 IEEE.  Personal use of this material is permitted.  Permission from IEEE must be obtained for all other uses, in any current or future media, including }}}}

\AddToShipoutPicture{\put(590,740){\rotatebox{-90}{\scalebox{1}{\color{gray}reprinting/republishing this material for advertising or promotional purposes, creating new collective works, for resale or redistribution to servers or lists,}}}}

\AddToShipoutPicture{\put(580,740){\rotatebox{-90}{\scalebox{1}{\color{gray}or reuse of any copyrighted component of this work in other works.}}}}

\begin{document}

%
\title{Revisiting Binary Local Image Description for Resource Limited Devices}
%
%
%
\author{Iago Su\'arez$^{1, 3}$, Jos\'e M. Buenaposada$^{2}$ and Luis Baumela$^{1}$
\thanks{Manuscript received: March, 23, 2021; Revised May, 24, 2021; Accepted August, 9, 2021.}%
\thanks{This paper was recommended for publication by Editor C. Cadena upon evaluation of the Associate Editor and Reviewers' comments.}%
\thanks{This work was supported by Doctorado Industrial grant DI-16-08966 and MINECO project TIN2016-75982-C2-2-R.}
\thanks{$^{1}$Departamento de Inteligencia Artificial, Universidad Polit{\'e}cnica de Madrid, Campus Montegancedo s/n. 28660 Boadilla del Monte, Spain 
{\tt\small iago.suarez.canosa@alumnos.upm.es}, {\tt\small lbaumela@fi.upm.es}}%
\thanks{$^{2}$ETSII, Universidad Rey Juan Carlos, C/ Tulip{\'a}n, s/n. 28933 M{\'o}stoles, Spain 
{\tt\small josemiguel.buenaposada@urjc.es}}
\thanks{$^{3}$The Graffter. S.L. }%
\thanks{Digital Object Identifier (DOI): see top of this page.}
}
%
%

\markboth{IEEE Robotics and Automation Letters. Preprint Version. Accepted August, 2021}
{Su\'arez \MakeLowercase{\textit{et al.}}: Revisiting binary local image description for resource limited devices} 

%



\maketitle

\begin{abstract}
The advent of a panoply of resource limited devices opens up new challenges in the design of computer vision algorithms with a clear compromise between accuracy and computational requirements. 
In this paper we present new binary image descriptors that emerge from the application of triplet ranking loss, hard negative mining and anchor swapping to traditional features based on pixel differences and image gradients.
These descriptors, BAD (Box Average Difference) and HashSIFT,
establish new operating points in the state-of-the-art's accuracy vs.\ resources trade-off curve.
In our experiments we evaluate the accuracy, execution time and energy consumption of the proposed descriptors. We show that 
BAD bears the fastest descriptor implementation in the literature while
HashSIFT approaches in accuracy that of the top deep learning-based descriptors, being computationally more efficient. We have made the source code public\footnote{\texttt{\url{https://github.com/iago-suarez/efficient-descriptors}}}.
\end{abstract}


%
\IEEEpeerreviewmaketitle

\section{Introduction}
%
%
%
%
\IEEEPARstart{L}{ocal} image description represents salient image regions in a invariant way to unwanted image transformations such as viewpoint changes, lighting conditions blur or camera noise. This is a basic step in the pipeline of computer vision applications in robotics~\cite{galvez2012bags,campos2020orb,GomezOjeda2019plslam,schlegel2018hbst}. 
Performing this process efficiently, fast and with low energy consumption, is also important for the quality of the final application. However there is a trade-off between efficiency and the robustness of the descriptor, because the simpler the descriptor the less invariant to the image transformations.

Local descriptors are typically vectors of floating point~\cite{lowe2004distinctive, Balntas2016tfeat, tian2017l2net, mishchuk2017working} or binary~\cite{calonder2010brief, rublee2011orb, strecha2012ldahash, yang2014ldb, balntas2015bold, trzcinski2015learning, levi2016latch, ye2019cdbin} values that are invariant to many image transformations.
Binary descriptors, the subject of this paper, are widely used in the context of resource limited devices, given their superior matching efficiency. The most accurate approaches are based on Deep Neural Nets (DNN)~\cite{ye2019cdbin}, that, in practical situations, provide a meager gain in accuracy at the price of a sharp increase in computational requirements~\cite{schonberger2017comparative}. Hence, in a real application, it is fundamental to balance the accuracy and computational requirements of the adopted solution. This is the reason why conventional descriptors such as ORB~\cite{rublee2011orb}  and SIFT~\cite{lowe2004distinctive,arandjelovic2012three} are still widely used in practical applications. The former because it is computational efficiency, and the latter because of the compromise between accuracy and efficiency.

\begin{figure}
	\centering
	\includegraphics[width=\linewidth]{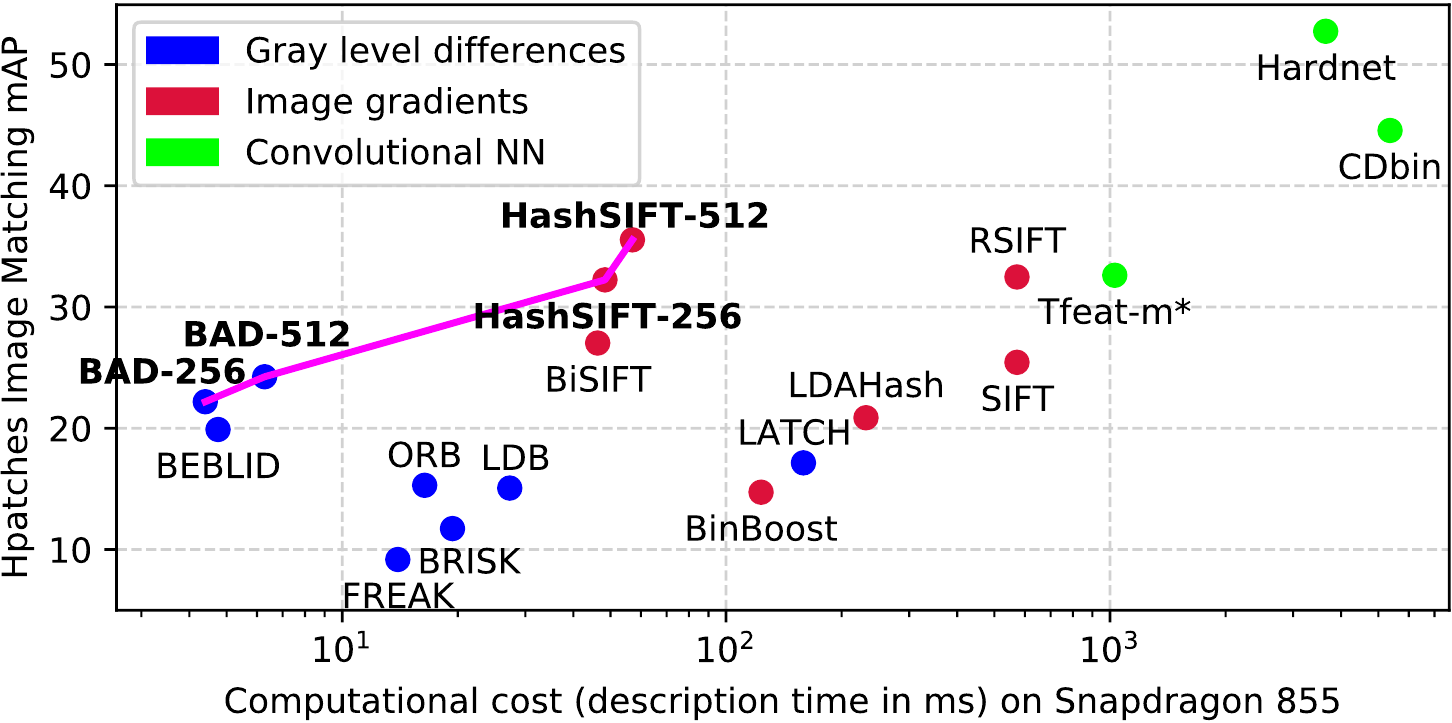}
	\caption{Accuracy vs.\ computational cost trade-off curve on CPU. BAD and HashSIFT provide new operating points on the top left region, crucial for a resource limited device.}
	\label{fig:accuracy_vs_efficiency}
\end{figure}

Our goal in this paper is to provide the robotics community with new 
computationally efficient binary descriptors
(see Fig.~\ref{fig:accuracy_vs_efficiency}). 
To this end we revisit efficient traditional local image features in combination with recent descriptor learning techniques.
We borrow from the Deep Learning (DL) literature learning techniques such as Triplet Ranking Loss (TRL) and  Hard Negative Mining (HNM) with anchor swap. The former brings different descriptions of the same scene point closer~\cite{Balntas2016tfeat}. The latter challenges the TRL with different scene points that have the closest description~\cite{mishchuk2017working}.

We apply these techniques to descriptors based on pixel differences. To learn its parameters, we propose a novel feature selection approach that efficiently searches for the best Box Average Difference (BAD) of two pixels by optimizing the TRL. The resulting BAD descriptors significantly outperform ORB's accuracy (see Fig.~\ref{fig:accuracy_vs_efficiency}).
We also revisit the binarization of SIFT by learning a linear hashing projection that optimizes the TRL. We name this more accurate binary descriptor HashSIFT.

In our experiments we evaluate the accuracy, execution time and energy consumption of the proposed descriptors. We show that they are the most accurate when confronted with competing techniques with similar computational requirements (see Fig.~\ref{fig:accuracy_vs_efficiency}).
They provide new operation points on the accuracy vs. resources trade-off curve, a useful tool when running computer vision applications with a tight energy budget. Using the most efficient BAD implementation we increase in more than 3 points the mAP in solving a planar image registration problem with OpenCV's ORB and reduce in about 30\% the estimation time. For the same problem, HashSIFT performs on par with the top DL descriptors, being more efficient.

In summary, in this paper we use recent optimization techniques based on TRL with HNM and anchor swap to contribute with
\begin{itemize}
    \item BAD, a fast binary descriptor based on pixel differences. To this end we propose an efficient algorithm for searching in the space of scales of pixel values (box size) and intensity biases.
	\item HashSIFT, a binary descriptor based on image gradients, with better accuracy than SIFT.
\end{itemize}

\section{Related work}

The computational performance of a local descriptor depends to a great extent on the underlying local image feature. Fig.~\ref{fig:accuracy_vs_efficiency} shows the existence of three clusters around roughly $10^1$, $10^2$ and $10^3$ milliseconds per frame (mspf) description time, depending respectively on whether the approach uses pixel gray value differences, image gradients or convolutional features. We organize our literature review around this key feature.

The fastest local features are based on differences of pixel gray values. BRIEF~\cite{calonder2010brief}, ORB~\cite{rublee2011orb}, LDB~\cite{yang2014ldb}, BOLD~\cite{balntas2015bold}, LATCH~\cite{levi2016latch}, FREAK~\cite{alahi2012freak}, BRISK~\cite{leutenegger2011brisk} and BEBLID~\cite{suarez2020beblid} are the most efficient in the literature. 
BRISK uses a fixed grid of binary tests, while BRIEF samples a random pattern, around the detected interest point. Other approaches use a greedy procedure to uncorrelate their features (e.g. ORB, FREAK, LATCH), whereas BOLD uses an online procedure to select the most robust features for each patch.
LDB~\cite{yang2014ldb} compares intensities and gradients obtained from the integral image.
ORB is still extensively used~\cite{galvez2012bags,campos2020orb,GomezOjeda2019plslam} as a local image descriptor. 
This is because its efficiency enables the processing of images in very few milliseconds on a CPU and the gains in accuracy achieved by more recent approaches do not pay off the computational cost they require.
Alternative and computationally more expensive binary descriptors are based on pixel gradients, OptConv~\cite{simonyan2014learning}, BinBoost~\cite{trzcinski2015learning}, and RFD~\cite{fan2014rfd}. 
OptConv selects its features by optimizing a loss based on a quartet of patches. BinBoost, LDB and BEBLID use pairs of similar and dissimilar patches to solve a classification problem with an exponential loss.
The TRL used in BAD outperforms the pairwise loss because it is closely related to the image matching problem. 

Descriptors using image gradient features are roughly one order of magnitude slower, but represent a good compromise between accuracy and computational efficiency. SIFT~\cite{lowe2004distinctive} is the most representative descriptor in this group. Although it was introduced more than twenty years ago, in practical situations it has remarkably
stood the test of time~\cite{schonberger2017comparative}. SURF~\cite{bay2006surf} approximates the first and second order derivatives with the integral image, that we also use to speed up our box differences. 
Both approaches produce a descriptor based on a histogram of gradient orientations. A key performance improvement for histogram-based representations is using the Hellinger distance to produce the so-called RootSIFT descriptor~\cite{arandjelovic2012three}.
%
%
BiSIFT~\cite{bellavia2020bisift} binarizes SIFT with a hand-crafted set of bin comparisons, inspired by BisGLOH2~\cite{bellavia2018rethinking}. In HashSIFT we also binarize SIFT by learning a linear projection that optimizes a TRL with HNM and anchor swap, that provides a better solution.
LDAHash~\cite{strecha2012ldahash} was an earlier attempt with the same aim. 
In~\cite{norouzi2012hamming} they propose a hashing approach based on metric learning in the hamming space. They show the superiority of TRL over pairwise losses. 
FastHash~\cite{lin2015supervised} uses graph cuts to determine the binary codes and model their prediction as a classification problem.
In BAD, we also use a greedy feature selection algorithm like FastHash. Instead of extracting a large pool of features, we only use grayscale differences thresholded to obtain the descriptor bits.
Further, we let the optimization choose the best hash codes to represent the data, instead of forcing the learner to use a predefined set.


DL models produce binary descriptors with the best accuracy, but with a high computation cost. 
L2Net~\cite{tian2017l2net} is a well-known approach using a fully convolutional architecture.
Tfeat~\cite{Balntas2016tfeat} proposes a simpler convolutional neuronal network with 2 layers and 64 channels trained using anchor swap negative mining.
Changing the sampling strategy by a hardest in batch, Hardnet~\cite{mishchuk2017working} achieves even better results with an L2Net architecture. DOAP~\cite{he2018local} changes the loss to directly optimize the mean Average Precision (mAP).
Other methods such as DCBD-MQ~\cite{duan2019dbdmq}, DeepBit~\cite{Lin2019DeepBit} or GraphBit~\cite{duan2018GraphBit} learn the binary codes in an unsupervised way.
The state of the art in binary description is CDBin~\cite{ye2019cdbin}. It uses five convolutional layers with 128 channels to optimize a TRL with extra restrictions in the correlation, quantization and distribution of the resulting bits. 
Despite the prohibitive cost of deep models in mobile devices, some of the ideas used to improve their performance may also be used in the context of more efficient traditional descriptors~\cite{buenaposada2021}. Our proposal in this paper follows this line of action.

\section{Learning efficient local descriptors}

In this section we describe the methodological foundation of our proposed descriptors.

\subsection{Box Average Difference descriptor (BAD)}
\label{sec:our_feature_selection}


We introduce a greedy procedure to select an uncorrelated set of pixel differences that discriminate between similar and different image patches. 
Our binary weak-descriptor $h(\vx ; f, \theta)=\{+1 \mbox{ if } f(\mathbf{x}) \leq \theta; -1 \mbox{ otherwise}\}$ is a decision stump on $f(\vx; \vu_1, \vu_2, s)$, a feature computed from the
difference of the average gray values of two boxes of size $s$ centered in pixels $\vu_i$ of patch $\vx$. It approximates the image gradient at a given direction and scale, and it can be efficiently computed using integral images~\cite{suarez2020beblid}. 
Learning the size of the box lets the feature adapt to the best scale to describe the patch. By selecting the optimal threshold, $\theta$, we drive the average feature value to zero. This, together with the TRL introduced next, will enable us to binarize our descriptor by just zero thresholding.


Our goal is to select from a very large pool of candidates features a subset that discriminates between same and different image patches. The selected weak-descriptors can be arranged in a vector $\vh(\vx) = \left[h_1(\vx), \cdots, h_K(\vx)\right]^\top$ resulting in a binary descriptor. 
Let $\cS(\vx, \vy) = \vh(\vx)^\top\vh(\vy)$
be the similarity between the binary descriptors of patches $\vx$ and $\vy$. $\cS(\vx, \vy)$ is positive if most weak-descriptors for both patches have the same sign and negative otherwise.

Let $\cT = \{(\va_i, \vp_i, \vn_i)\}_{i=1}^N$ be a set of $N$ patch triplets. 
All patches have been re-scaled to the same size, 32$\times$32 pixels, where $\va_i$ is the anchor patch, $\vp_i$ is a positive (i.e. same scene point as the anchor) and $\vn_i$ is a negative patch w.r.t. $\va_i$ (i.e. different scene point). We use $\cT$ to find $K$ features that minimize 
\begin{equation}
\cL_{\text{fs}} = 
\sum_{i=1}^N \left[\tau - \cS(\va_i,\vp_i) + \cS(\va_i,\vn_i) \right]_+,
\label{eq:loss-function-feature-selection}
\end{equation}
where $[\cdot]_+ = \max(0, \cdot)$ is the TRL with margin $\tau$. 

We minimize Eq.~\ref{eq:loss-function-feature-selection} incrementally in a greedy fashion where the $t$-th weak-descriptor is chosen taking into account the $t-1$ previously selected. Thus, the optimization problem we solve is given by
$h_t^* = \arg\min_{h} \cL_{h}$,
\begin{equation}
\begin{split}
\cL_{h}(\cT, \theta) = \sum_{i=1}^N \left[\tau -\cS^{(t-1)}(\va_i,\vp_i) - h_t(\va_i)h_t(\vp_i) \right. \\
\left. + \cS^{(t-1)}(\va_i,\vn_i) + h_t(\va_i)h_t(\vn_i)\right]_+,
\end{split}
\label{eq:optimization-per-step}
\end{equation}
where $\cS^{(t-1)}(\vx,\vy)$ 
is the similarity computed with the first $t-1$ elements. 
\begin{algorithm}
	\caption{Feature selection algorithm}
	\label{alg:feature-selection}
	\textbf{Input:} $\cP$, patches with 3D point class label\\ 
	\textbf{Output:} $\vh = [h_1, ..., h_K]$ 
	\begin{algorithmic}[1]
		\STATE $\vh := \emptyset$
		\FORALL{$k = 1,...,K$}
		\STATE $\cT := \text{sampleTriplets}(\cP, N)$ \label{alg:sampleTriplets}
		\STATE $ \cF := \text{sampleFeatures}()$ \label{alg:sampleFeatures}
		\STATE $\cL^* := \infty, \theta^* := -\infty$
		\FORALL {$j = 1,..., J$}
		\STATE $\theta_j, \cL_j := \text{findThreshold}(\cT, f_j(\cdot), \cL_{h}(\cdot))$
		\IF{$\cL_j < \cL^*$}
		\STATE $\cL^*$ := $\cL_j$; $f^*$ := $f_j$; $\theta^*$ := $\theta_j$
		\ENDIF
		\ENDFOR
		\STATE $h_k := \{+1 \mbox{ if } f^*(\mathbf{x}) \leq \theta^*; -1 \mbox{ otherwise}\}$ 
		\STATE $\vh := \vh \cup h_k$ \\
		\ENDFOR
		\RETURN $\vh$
	\end{algorithmic}
\end{algorithm}

In Algorithm~\ref{alg:feature-selection} we show the steps involved in each iteration of the optimization. The input is the set of training image patches and labels, $\cP=\{(\vx_i, l_i)\}_{i=1}^{M}$, where label $l_i$ is the 3D point index of $\vx_i$. 
At the beginning of each iteration the set of training patches, $\cP$, is sampled to form a set of triplets, $\cT = \{(\va_i, \vp_i, \vn_i)\}_{i=1}^{N}$ using HNM~\cite{mishchuk2017working} and anchor swap~\cite{Balntas2016tfeat}. We also sample a random set of ($J = 1000$) candidate features at each iteration, $\cF=\{f_j(\vx)\}_{i=1}^J$, from which we select the feature with the lowest $\cL_{h}$. 
The scale selection is done implicitly because the function \emph{sampleFeatures} in line \ref{alg:sampleFeatures} of Alg. \ref{alg:feature-selection} samples boxes at all possible scales.

This feature selection method does not force each bit to have zero mean and to be uncorrelated. 
However, these properties emerge when we optimize Eq.~\ref{eq:optimization-per-step}, since for most weak-descriptors the largest loss decrease is achieved by evenly assigning -1 and 1 values to the triplet samples so that $\mathbb{E}_{\vx \in \cT}(h_t(\vx))\approx 0$.

\begin{figure}[t]
	\begin{center}
		\includegraphics[width=1.0\linewidth]{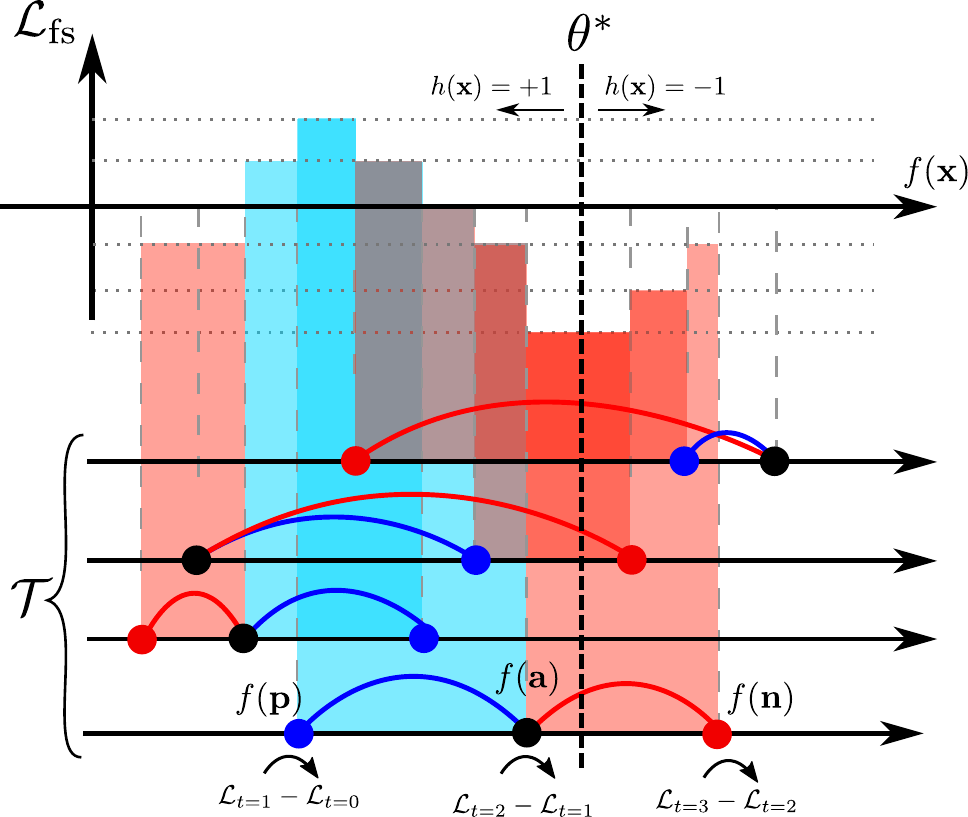}
	\end{center}
	\caption{Space-splitting to find the optimal feature threshold for a TRL.
		Each point represents a patch. The anchor $\va$ is shown in black, the positive $\vp$ in blue and the negative $\vn$ in red. The loss decreases when the threshold severs a red tie and increases for the blue ones. The optimal threshold $\theta^*$ is the one that minimizes the accumulated loss.
	}
	\label{fig:best-th-alg}
\end{figure}

Given a candidate feature, $f(\vx)$, a critical step in the algorithm is selecting the threshold, $\theta$, with the lowest $\cL_h$. To this end, in Algorithm~\ref{alg:threshold-rate}, we introduce a procedure to efficiently estimate the optimal $\theta$. Our method is a generalization of that in~\cite{shakhnarovich2005learning} for any triplet-based loss. It sorts the feature values, $f(\vx)$, for all patches in $\cT$ and shifts the threshold from the lowest to the highest value. As we show in Fig.~\ref{fig:best-th-alg}, the loss changes only when $\theta$ goes beyond any patch $f(\vx)$ value. Thus, at that point we can compute the change (increase or decrease) in $\cL_h$. 
If we initialize the accumulated loss with the value achieved with the smallest threshold, the minimum of the cumulative sum of loss changes $\Delta\cL$, represented in the upper part of Fig.~\ref{fig:best-th-alg} with the color bars (the redder the lower the loss function and the bluer the higher), will produce the best $\theta$.

Let $P$ be the number of patches in $\cT$. Brute force search would threshold candidates between each pair of consecutive feature values 
and iterate over all the patches to compute $\cL_h$, resulting in $O(P^2)$ complexity. On the other hand, the sorting in line \ref{alg:sort-patches} of Algorithm \ref{alg:threshold-rate} dominates the complexity, being $O(P\log P)$. If we fix a minimum precision for $\theta^*$ we can speed up this process using integer sorting, that is $O(P+m)$, where $m$ is the size of the search region divided between the desired precision. For example, with BAD, the search space is $[-255, 255]$, so for a precision of 0.1 we have $m=\frac{511}{0.1} = 5110$ possible values for $\theta$. This speed-up is critical to search among thousands of feature candidates in $\cF$ each iteration. 

\begin{algorithm}
	\caption{$\text{findThreshold}(\cT, f, \cL)$}
	\label{alg:threshold-rate}
	\textbf{Input:} $\cT=\{(\va_i, \vp_i, \vn_i)\}_{i=1}^{N}$\\
	\textbf{Input:} A feature extraction function, $f$ \\
	\textbf{Input:} A loss function, $\cL$ \\ 
	\textbf{Output:} $\theta^*, \cL^*$ 
	\begin{algorithmic}[1]
		\STATE $v$ := $\emptyset$
		\FORALL{$i = 1,...,N$}
		\STATE $(\vz^{(1)}, \vz^{(2)}, \vz^{(3)})$ := $(\va_i, \vp_i, \vn_i)$
		\FORALL{$m = 1,2,3$}
		\STATE $\cL^-$ :=$\cL(\{(\va_i, \vp_i, \vn_i)\}; \theta=f(\vz^{(m)}) - \epsilon)$
		\STATE $\cL^+$ :=$\cL(\{(\va_i, \vp_i, \vn_i)\}; \theta=f(\vz^{(m)}) + \epsilon)$
		\STATE $\Delta\mathcal{L}_m =\cL^+ - \cL^-$
		\STATE $v$ := $v \cup \left\{(f(\vz^{(m)}), \Delta\mathcal{L}_m)\right\}$.
		\ENDFOR
		\ENDFOR
		\STATE $v_s$ := sortByFeatureValue($v$) \label{alg:sort-patches}
		\STATE $\cL^*$ := $-\infty$, $\theta^*$ := $-\infty$
		\STATE $\cL_{\text{sum}}$ := $\cL(\cT; \theta=-\infty)$
		\FORALL{$j = 1,...,P$}
		\STATE $(f_j, \Delta\cL_j)$ := $v_s(j)$
		\STATE $\cL_{\text{sum}}$ := $\cL_{\text{sum}} + \Delta\cL_j$
		\IF {$\cL_{\text{sum}} < \cL^*$}
		\STATE $\cL^*$ := $\cL_{\text{sum}}$, $\theta^*$ := $f_j + \epsilon$
		\ENDIF
		\ENDFOR
		\RETURN $\theta^*, \mathcal{L}^*$
	\end{algorithmic}
\end{algorithm}

\subsection{Binarizing SIFT}
\label{sec:our_binarization}
If more computational power is available, we can use gradient orientations instead of grayscale differences. The well-known SIFT features are powerful, but also highly correlated and with non-zero mean, since it is a histogram.
Thus, in this section we devise a method to learn a linear hashing method that fixes these issues before binarization.

Let $\vf(\vx) = \left[f_1(\vx), \cdots, f_K(\vx)\right]^\top$ be the vector of real valued features provided by SIFT. 
We estimate matrix $\mB$
to produce the binary descriptor 
%
%
$\vD(\vx) = \text{sgn}(\mB \left[\vf(\vx)^\top, 1\right]^\top)$.
%
For training, we approximate 
$\tilde{\vD}(\vx) = \text{tanh}( \mB \left[\vf(\vx)^\top, 1\right]^\top)$ and find $\mB=\arg\min_{\mB} \cL_B$, where
\begin{equation}
\cL_B = \sum_{i=1}^{N} \left[\tau - \tilde{\vD}(\va_i)^\top \tilde{\vD}(\vp_i) + \tilde{\vD}(\va_i)^\top \tilde{\vD}(\vn_i) \right]_+.
\label{eq:hash-loss-func}
\end{equation}

We minimize $\cL_B$ with stochastic gradient descent using Adam with learning rate 0.0002. We randomly initialize the elements of $\mB$ with a Gaussian distribution, $\mathcal{N}(0, 0.25)$. 
At each iteration $\cP$ is sampled randomly to form a batch of triplets, $\cT = \{(\va_i, \vp_i, \vn_i)\}_{i=1}^{N}$, using HNM and anchor swap. 
To prevent over-fitting we use data augmentation adding a small random rotation, scale, illumination change, blurring and pixel Gaussian noise to each patch.
Since this optimization process is a continuous version of that presented in section \ref{sec:our_feature_selection}, 
$\mB$ transforms SIFT into a set of uncorrelated values with a median close to 0, so it can be trivially binarized.

\section{Experiments}

We evaluate our descriptors with the
Brown~\cite{brown2011dataset}, Oxford~\cite{mikolajczyk2005performance}
and Hpatches~\cite{balntas2017hpatches} data sets and on a real Structure from Motion (SfM) 
problem~\cite{schonberger2017comparative}. 
We also compare the execution time on a laptop and two smartphone CPUs following \cite{suarez2020beblid}. We train our descriptors with the \texttt{Liberty} data set~\cite{brown2011dataset} as is common in the literature.
For LDAHash, Tfeat-m*, CDbin, DOAP, LDB, BiSIFT and BEBLID we use the implementation provided by the authors, for the other methods we use OpenCV~\cite{opencv_library}.
Our descriptors are implemented in C++ with Python wrappers.
%

We perform our evaluation in two dimensions: description accuracy and computational efficiency. To have a fair comparison in this domain we organize the experimentation in groups, depending on the underlying image feature, since, as discussed in the previous section, it determines the computational complexity of the descriptor. The first group clusters techniques based on pixel differences. The second group includes approaches using image gradients. Finally, in the third group we represent DL based techniques. 
Tfeat-m* is chosen because it is the simplest and most efficient model, Hardnet and DOAP respectively as representative top floating point and binary descriptors, 
and CDbin because it is the most accurate binary descriptor.
Within each group \first{first}, \second{second} and \third{third} ranked results are shown respectively in blue, green and red colors.

\subsection{Ablation Study}
\label{sec:ablation_study}

Some efficient detectors, such as BinBoost, LATCH or BEBLID, are trained with a binary classification loss defined over pairs of patches.
However, in the area of DL, there is solid recent evidence in favor of the TRL~\cite{Balntas2016tfeat}. In our analysis we also include the use of binary loss to confirm that for our descriptors the TRL also obtains better results.

Table~\ref{tab:bad_ablation} shows the contribution of each component in the design of BAD. As a baseline we select ORB~\cite{rublee2011orb},
a descriptor based on the difference of pairs of pixel values after smoothing the image with a 5$\times$5 box filter. These features are equivalent to BAD features with $s=5$ and $\theta=0$. 
This baseline reaches 15.30 mAP in the Hpatches image matching task. By changing the feature selection to a binary classification loss the mAP improves in 1.35 mAP points. If we include a learned threshold, $\theta$, we get a significant improvement of 3.3 points in mAP. If we use the TRL we get 1.72 additional points. Learning the feature scale 
the mAP increases again in 0.28 points. Finally applying data augmentation we get an additional 0.24 points increase. In summary, the most important components of our methodology are learning the feature threshold, $\theta$, that increases the accuracy and drives to zero its mean value, and the TRL, that models the asymmetric nature of the matching problem. 
%
Note that binary classification loss (with AdaBoost) and learned threshold are the key elements that were used in BEBLID. Thus, when changing the loss to TRL and adding anchor swap and box data augmentation, we improve the results of BEBLID from 19.95 to 22.19 mAP. This is a significant improvement and makes clear the difference between BEBLID and BAD.

\begin{table}
	\centering
	\caption{BAD ablation study in the Image Matching task of Hpatches~\cite{balntas2017hpatches}.}
	\begin{tabular}{|c|c|c|c|c|c|c|}
		\hline
		ORB &
		\begin{tabular}[c]{@{}c@{}}Binary \\ classif. \\ loss\end{tabular} &
		\begin{tabular}[c]{@{}c@{}}Learned \\ Thr ($\theta$)\end{tabular} &
		\begin{tabular}[c]{@{}c@{}}Triplet \\ Ranking \\ Loss \end{tabular} &
		\begin{tabular}[c]{@{}c@{}}Box Scale \\ selection\end{tabular} &
		\begin{tabular}[c]{@{}c@{}}Data \\ Aug \end{tabular} &
		\begin{tabular}[c]{@{}c@{}}Matching \\ mAP\end{tabular} \\ \hline
		\checkmark &   &   &   &   &   & 15.30 \\ \hline
		& \checkmark &   &   &   &   & 16.65 \\ \hline
		& \checkmark & \checkmark &   &   &   & 19.95 \\ \hline
		&   & \checkmark & \checkmark &   &   & 21.67 \\ \hline
		&   & \checkmark & \checkmark & \checkmark &   & 21.95 \\ \hline
		&   & \checkmark & \checkmark & \checkmark & \checkmark & 22.19 \\ 
		\hline
	\end{tabular}
	\label{tab:bad_ablation}
\end{table}

In Table~\ref{tab:hashsift_ablation} we show the ablation study for HashSIFT. In this case if we optimize the typical binary classification loss we get 28.64 mAP. With a TRL and HNM we get an improvement of 2.34 points in mAP. Finally, data augmentation gives us 1.26 additional points. This also confirms that, for hashing, the TRL with HNM is the key ingredient to achieve top accuracy.  

\begin{table}
	\centering
	\caption{HashSIFT ablation study in HPatches Image Matching task~\cite{balntas2017hpatches}.}
	\begin{tabular}{|c|c|c|c|c|c|}
		\hline
		\begin{tabular}[c]{@{}c@{}}Binary \\ classif. \\ loss \end{tabular} & 
		\begin{tabular}[c]{@{}c@{}}Triplet \\ Ranking \\ Loss\end{tabular} &
		\begin{tabular}[c]{@{}c@{}}HNM\end{tabular} &
		\begin{tabular}[c]{@{}c@{}}Data \\ Aug\end{tabular} &
		\begin{tabular}[c]{@{}c@{}}Matching \\ mAP\end{tabular} \\ \hline
		\checkmark &   &   &   &  28.64 \\ \hline
		& \checkmark & \checkmark &   & 30.98 \\ \hline
		& \checkmark & \checkmark & \checkmark &  32.24 \\ 
		\hline
	\end{tabular}
	\label{tab:hashsift_ablation}
\end{table}

\begin{table}
	\centering
	\caption{Results on the Brown (third and fourth columns) and Oxford (last column) data sets. Patch verification performance on the Brown data set is measured with positive rate at 95\% of TPR and with mAP for the Oxford benchmark.} 
	\resizebox{\linewidth}{!}{%
		\begin{tabular}{|l|c||c|c||c|} \hline
			\multirow{3}{*}{\textbf{Method}} & \multirow{3}{*}{\textbf{Size}} &  \multicolumn{2}{c||}{\textbf{FPR-95}} & \textbf{mAP} \\
			& &  \multicolumn{2}{c||}{\textbf{Liberty (train)}} & \textbf{Liberty (train)} \\ 
			& & \textbf{Notredame (test)} & \textbf{Yosemite (test)} & \textbf{Oxford (test)} \\ 
			\hline
			BRISK~\cite{rublee2011orb}            & 512b & 69.82 & 68.40 & 32.25 \\ 
			ORB~\cite{rublee2011orb}              & 256b & 38.02 & \third{40.58} & 51.72\\ 
			LDB~\cite{yang2014ldb}                & 256b & 52.24 & 49.78 & 51.53 \\
			LATCH~\cite{levi2016latch}            & 512b & 47.04 & 47.05 & 52.48\\ 
			BEBLID~\cite{suarez2020beblid}        & 256b & \third{36.42} & 40.84 & \third{53.59}\\ 
			BAD (ours)                            & 256b & \first{30.97} & \second{35.67} & \second{55.06}\\ 
			BAD (ours)                            & 512b & \second{31.77} & \first{34.69} & \first{57.20} \\ 
			\hline 
			SIFT~\cite{lowe2004distinctive}       & 128f & 21.96 & 28.07 & 56.40 \\ 
			RSIFT~\cite{arandjelovic2012three}    & 128f & 23.79 & 32.16 & \first{64.13}\\ 
			BinBoost~\cite{arandjelovic2012three} & 256b & 15.76 & 22.79 & 44.00 \\
			ConvOpt~\cite{simonyan2014learning}   & 1024b & \second{8.25} & \third{14.84} & 52.34 \\ 
			BiSIFT~\cite{bellavia2020bisift}      & 482b & 17.94 & 20.82 & 57.70 \\
			LDAHash-DIF~\cite{strecha2012ldahash} & 128b & 41.12 & 38.93 & 53.34 \\
			HashSIFT (ours)                       & 256b & \third{9.49}  & \second{13.85} & \third{59.22}\\ 
			HashSIFT (ours)                       & 512b & \first{7.63}  & \first{10.96} & \second{62.21} \\ 
			\hline
			Tfeat-m*~\cite{Balntas2016tfeat}      & 128f & \third{3.12}  & \third{7.82} & \third{40.55}\\ 
			Hardnet~\cite{mishchuk2017working}    & 128f & \first{0.57}  & \first{2.13} & \first{63.27} \\ 
			CDbin~\cite{ye2019cdbin}              & 256b & \second{1.29}  & \second{4.43} & \second{53.51} \\ 
			\hline
		\end{tabular}
	}
    \label{tab:brown-oxford-results}
\end{table}

\subsection{Brown data set}
\label{sec:experiment-brown}
The Brown data set~\cite{brown2011dataset} contains patches detected with Difference of Gaussians (DOG) in three scenes: \texttt{Liberty}, \texttt{Notredame} and \texttt{Yosemite}. Patches are distributed into corresponding and non-corresponding pairs. The ability to solve this classification problem is evaluated using the False Positive Rate at 95\% of True Positive Rate (FPR-95).
This metric is good for patch verification but it is far from the optimal operation point for image matching problems, due to the large number of negatives and the cost of False Positives.

The third and fourth columns of Table~\ref{tab:brown-oxford-results} show the results on Brown. In the first group of approaches, based on pixel differences, both BAD incarnations, with 256 and 512 bits, achieve the best accuracy, with the lowest FPR-95.
%

In the second group of features in Table~\ref{tab:brown-oxford-results}, 
%
our hashing technique uncorrelates SIFT producing the best gradient-based binary 
descriptor, outperforming the floating-point RootSIFT by a large margin in the patch verification problem as well as the other SIFT binarizations.  Here LDAHash has the worst results because it was tuned for patch retrieval and image matching. 

Finally, in the third group, the best DL binary descriptor is CDBin, but at the price of spending much more computation and energy (see Table~\ref{tab:execution_times}). Here, however, the performance of the best floating-point descriptor, Hardnet, is significantly better.

\subsection{Oxford data set}
\label{sec:oxford-experiment}
The Oxford data set~\cite{mikolajczyk2005performance} contains images of 8 planar scenes with varying perspective, blurring, scale, rotation, illumination and JPEG compression level. 
We use OpenCV's SIFT (DoG) detector and crop each feature in a patch of width $6.75$ times the diameter of the detected region. We describe each region and match them using Nearest-Neighbors. To compute the mean Average Precision (mAP) we produce a Precision-Recall curve varying the matching distance threshold between the first image of each sequence and the rest (see fifth column of Table~\ref{tab:brown-oxford-results}).

BAD beats again all its competitors in the first group by a significant margin.
Within the gradient based descriptors group, RootSIFT (RSIFT) has the best accuracy, however with a floating-point descriptor. HashSIFT-512 (bits) and HashSIFT-256 (bits) are in second and third positions. 
Finally, concerning DL descriptors we can see that, for this planar matching problem where the appearance change of points is limited, the extra learning capacity provided by DL is useless and gradient based
descriptors are on par with them (HardNet 63\% mAP  vs. RSIFT 64.13\% or HashSIFT-512 62.21\%). However,  DL methods are at least one order of magnitude slower (see Sec. \ref{sec:experiment-times} and Fig.~\ref{fig:accuracy_vs_efficiency}). 

\subsection{Hpatches}
\begin{figure*}
	\begin{center}
		\includegraphics[width=1.0\textwidth]{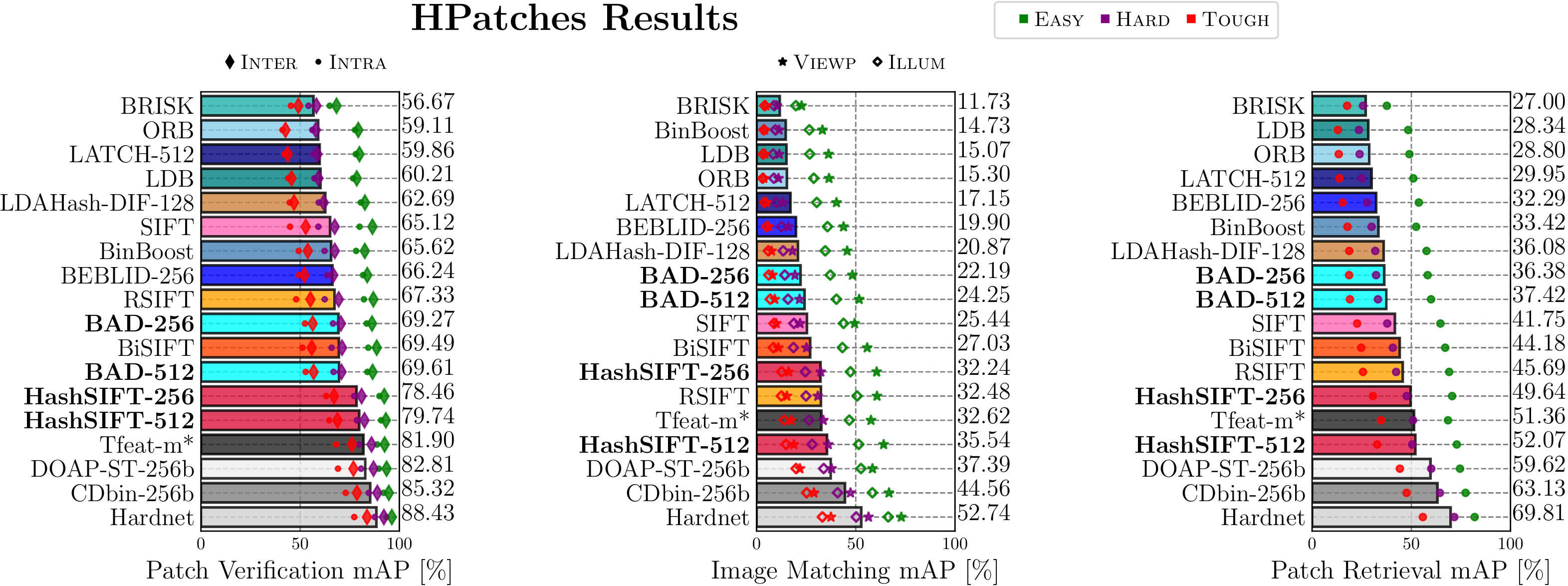}
	\end{center}
	\caption{Results in the full split of the Hpatches benchmark. We show methods based in pixel differences in blue colors, the methods based on gradients in red and orange colors and the DL methods in gray.
	}
	\label{fig:hpatches-results}
\end{figure*}

Hpatches~\cite{balntas2017hpatches} 
evaluates the mAP in three different tasks: patch verification, image matching and patch retrieval. 
In Fig. \ref{fig:hpatches-results} we show the results for the three descriptor groups.  Methods based on pixel differences are displayed in shades of blue, image gradients in red-orange and  DL ones in gray. 

In the verification problem, with 80\% negatives and 20\%  positives, BAD and HashSIFT are ranked right after DL ones.
BAD-512 improves not only all the efficient methods but also the gradient-based ones.

In the matching problem, if we consider the binary descriptors, CDBin, 44.56\% mAP, is the top performer. The third is HashSIFT-512, 35.54\% mAP, not far  from DOAP and outperforming Tfeat-m* despite its convolutional and floating point nature. 
This result makes HashSIFT-512 a good alternative to RSIFT in matching because of its binary nature and superior accuracy. 
In the first group, BAD outperforms all the other methods, specifically BAD-256 performs 45\% better than the well known ORB. 

In the patch retrieval problem descriptor algorithms behave similarly to matching.
In summary, in Hpatches, HashSIFT-512 is the best performing non-DL descriptor and should be preferred to RSIFT.
BAD establishes a new state-of-the-art among the most efficient descriptors in Hpatches.

\subsection{Reconstruction evaluation in the ETH benchmark}

The above tests assess descriptors' performance independently of the target application. In contrast the ETH benchmark~\cite{schonberger2017comparative} uses a full reconstruction pipeline to carry out the evaluation.

In Table~\ref{tab:eth-reconstruction-results} we show the results in the ETH data set for one complex scene, \emph{Madrid Metropolis},
which has 1,344 images and none of the algorithms register them all. Thus, the matching robustness of the descriptor can be studied by the number of reconstructed sparse points (\emph{\#Sparse Pts}).

In such a very complex scene, BAD-512 and BAD-256 reconstruct respectively 39\% and 41\% more points than ORB, showing the superiority of our features and training process.
On the other hand, HashSIFT-512 and HashSIFT-256 respectively reconstruct 6.5\% and 4.3\% more sparse points than RootSIFT. 
The number of registered images (\emph{\#Registered}) is also important, BAD-512 registers 622 while ORB only 457. On the other hand, both HashSIFT descriptors register 720 images while RootSIFT 729. These results confirm that, even in the most complex scenes, HashSIFT can be used as a drop-in replacement of RootSIFT with the additional advantage of being binary.
Moreover, in this scene HashSIFT reconstructs more sparse points than CDBin, the best DL binary  descriptor. HashSIFT-512 provides  17\% more reconstructed points than CDBin but with less registered images (720 vs 769 of CDBin) due to its shallow nature.

In a real SfM problem, the best reconstruction with the largest number of points and registered images is obtained by the state-of-the-art DL descriptors. However, if they are to run in a mobile device or the size of the data set is extremely large, then DL methods are not a good option, as we will address in the next section. In this case, we have two possibilities. With a bigger computational budget, we use the best gradient based descriptor, HashSIFT-512, that reconstructs 17\% more sparse points than CDbin in \emph{Madrid Metropolis} but is able to register less images. Or, with lower computational budget we should use the 
efficient descriptor BAD-256 that reconstructs 26\% less points. %
We have also reconstructed the Fountain, Herzjesu and South Building obtaining similar results\footnote{More results can be found in the project website: \url{https://iago-suarez.com/efficient-descriptors}}. BAD-512 obtains always the best results among the descriptors based on gray-level pixel differences with 15741, 8220 and 148491 \emph{\#Sparse Pts} respectively (4.93\%, 7.89 \% and 7.89\% more than ORB). HashSIFT-512 also obtain the best results in its group with 16385, 8769 and 156888 \emph{\#Sparse Pts}. It shortens the gap with DL methods such as CDbin, 
 with 16607, 8997 and 160589 \emph{\#Sparse Pts}, due to the scenes simplicity (only 1.34\%,  2.53\% and 2.3\% less \emph{\#Sparse Pts} than CDBin).

\begin{table}
	\begin{center}
		\caption{Results for the ETH benchmark.}
		\resizebox{\linewidth}{!}{%
			\begin{tabular}{|lcccc|}
				\hline
				\textbf{} &
				\textbf{\# Registered} &
				\textbf{\# Sparse Pts} &
				\textbf{\# Obervations} &
				\textbf{Track Length} 
				\\ \hline
				\multicolumn{5}{|c|}{\textbf{Madrid Metropolis (1344 images)}} \\
				\hline
				ORB             & 457         & 135826          & 576138          & \second{4.241736} \\
				BEBLID-256      & 549         & 174257          & 705651          & 4.049484         \\
				LATCH           & \third{573} & \third{186886}  & \third{759581}  & 4.064408         \\
				BAD-256         & \second{600}& \first{192638}  & \second{789466} & \third{4.098184} \\
				BAD-512         & \first{622} & \second{189523} & \first{812243}  & \first{4.285723} \\
				\hline
				RSIFT           & \first{729} & \third{286519}   & \second{1136306} & \second{3.965901} \\
				Binboost        & 514         & 143622           & 629993           & \first{4.386466 } \\
				LDAHash-DIF     & \third{592} & 233862           & 804944           & 3.441961          \\
				HashSIFT-256    & \second{720} & \second{298920} & \third{1075450}  & 3.597785          \\
				HashSIFT-512    & \second{720} & \first{305237}  & \first{1160738}  & \third{3.802743}  \\
				\hline
				Tfeat-m*        & \third{690} & \second{262790} & \third{986470}   & \third{3.753834} \\
				HardNet         & \first{849} & \first{359610}  & \first{1438909}  & \second{4.001304} \\
				CDbin-256b      & \second{769} & \third{260690} & \second{1108018} & \first{4.250328} \\ \hline 
			\end{tabular}%
		}
	\end{center}
	\label{tab:eth-reconstruction-results}
\end{table}

\subsection{Execution time and energy consumption}
\label{sec:experiment-times}

In this section we run state-of-the-art binary descriptors in different platforms: a laptop with an Intel Core i7 8750H CPU, 12 cores and 16GB of RAM; a smartphone \textit{Samsung J5 2017} with an Exynox Octa S CPU, 8 cores and 2GB of RAM; a smartphone \textit{One Plus 7 Pro} with Snapdragon 855 CPU, 8 cores and 6GB of RAM. 

Table~\ref{tab:execution_times} contains the average time to process each ($\approx 800$ pixels wide) image in the Oxford~\cite{mikolajczyk2005performance} data set describing a maximum of 2000 keypoints per image. We also show in brackets the standard deviation in 5 executions.
\begin{table}
	\centering
	\caption{Average  five run image description time (ms). Also shown the standard deviation in parenthesis. "Size" reports the descriptor size in floating-point (f) or binary (b) values.}
	\footnotesize
	\resizebox{\linewidth}{!}{%
		\begin{tabular}{|c|c|c|c|c|}
			\hline
			\multirow{2}{*}{\textbf{Method}} & \multirow{2}{*}{\textbf{Size}} & \textbf{Intel Core i7} & \textbf{Exynox} & \textbf{Snapdragon} \\
			& & \textbf{8750H} & \textbf{Octa S}& \textbf{855} \\ 
			\hline
			BRISK    & 512b    & 14.94 ($\pm$0.31) & 164.75 ($\pm$4.10) & 19.38 ($\pm$0.19)  \\ 
			ORB      & 256b    & 12.07 ($\pm$0.33) & \third{100.04 ($\pm$1.16)} & 16.40 ($\pm$0.25) \\ 
			LDB      & 256b    & 17.48 ($\pm$0.8)  & 161.30	($\pm$4.44) & 27.32 ($\pm$0.11)\\ 
			BEBLID   & 256b    & \second{1.56 ($\pm$0.05)} & \first{20.04 ($\pm$0.31)} & \second{4.75 ($\pm$0.06)}  \\
			LATCH    & 512b    & 101.89 ($\pm$1.67) &  1509.66 ($\pm$24.35) & 159.02 ($\pm$0.24)  \\ 
			BAD      & 256b    & \first{1.53 ($\pm$0.04)} & \first{20.04 ($\pm$0.28)} & \first{4.40 ($\pm$0.06)}   \\ 
			BAD      & 512b    & \third{2.77 ($\pm$0.08)} & \second{31.63 ($\pm$0.33)} & \third{6.28 ($\pm$0.15)}  \\ 
			\hline \hline
			SIFT     & 128f    & 187.24 ($\pm$5.28) & 2519.86 ($\pm$21.41) & 572.30 ($\pm$5.70) \\ 
			BinBoost & 256b    & 84.65 ($\pm$ 2.28) & 786.80 ($\pm$60.82) & 123.33 ($\pm$ 1.73)\\ 
			BiSIFT   & 482b    & \first{30.16 ($\pm$ 0.77)} & \third{293.38 ($\pm$38.60)} & \first{46.29 ($\pm$3.02)} \\
			HashSIFT & 256b    & \second{31.41 ($\pm$ 0.73)} & \first{211.76 ($\pm$2.66)} & \second{48.36 ($\pm$ 1.32)} \\ 
			HashSIFT & 512b    & \third{33.80 ($\pm$ 1.67)} & \second{259.17 ($\pm0.62$)} & \third{57.35 ($\pm$ 0.42)} \\ 
			\hline \hline
			Tfeat-m* & 128f    & \first{332.05 ($\pm$ 7.41)}   & \first{4110.61 ($\pm$46.59)} &  \first{1028.94 ($\pm$59.25)} \\
			CDbin    & 256b    & \third{1184.82 ($\pm$ 26.24)} & \third{23219.06 ($\pm$ 274)} & \third{5363.37 ($\pm$383.79)} \\ 
			Hardnet  & 128f    & \second{763.08 ($\pm$ 15.27)}  & \second{15640.32 ($\pm$112.8)} & \second{3645.39 ($\pm$274.19)} \\
			\hline
		\end{tabular}
	}
	\label{tab:execution_times}
\end{table}
%
These results show that BAD bears the the fastest descriptor implementation in the state of art, marginally ahead of BEBLID. This is thanks to the use of very efficient image features computed with the integral image. 
FREAK is faster than ORB on smartphone CPUs, but it has the worst performance in Hpatches and ETH.
%
One order of magnitude slower, but still feasible for near real-time applications, HashSIFT, provides a good trade-off between accuracy and efficiency, being the fastest binary descriptor based on gradient features. Note that OpenCV's SIFT (SIFT in the table) computes a Gaussian pyramid for the entire image. Instead, our implementation of HashSIFT only describes a normalized $32\times32$ px image patch cropped around each local feature point. This results in a significant 6$\times$ description time speed up in a desktop and 11$\times$ in smartphone CPUs.

To ensure a fair comparison, we run all methods in CPU. On the smartphones we run CDBin in OpenCV's DNN module with Tengine\footnote{\url{https://github.com/OAID/Tengine}}  to speed up the execution time.
We consider the CPU run time as proxy of the computational resources required by DL methods, regardless of the hardware they run on.
CDBin, the state-of-the-art DL binary descriptor 
needs 100$\times$ more time than HashSIFT-256 in the smarphone CPUs and 1000$\times$ more than BAD-256. 
%

We also evaluate the energy consumption of a full image matching application in Table~\ref{tab:battery}. The top speed of $~30$FPS, the maximum frame rate allowed by the camera, is only achieved by the descriptors based on gray-level pixel differences. CPU time is counted separately for each CPU core, thus, BAD-256 has energy statistics similar to OpenCV's ORB but obtains faster total description times (Table \ref{tab:execution_times}) because it takes advantage of a  parallel implementation. 

HashSIFT is the most energy efficient among the second group of methods. Compared to OpenCV's SIFT implementation, our HashSIFT-256 implementation improvements compensate for its slightly higher complexity and make it three times faster.
DL methods drain the battery much faster, for example, CDBin requires 146$\times$ more energy per frame than the same application using BAD-256 and 62$\times$ more energy than HashSIFT-256, providing only a small gain in accuracy in a practical application (see Table~\ref{tab:eth-reconstruction-results}).

\begin{table}
	\centering
	\caption{Per frame statistics of a template matching application running for 10 minutes on a Snapdragon 855 CPU without networking and initial battery charge at 100\%.}
	\footnotesize
	\resizebox{\linewidth}{!}{%
		\begin{tabular}{|c|c|c|c|c|c|}
			\hline
			\multirow{2}{*}{\textbf{Method}} & \multirow{2}{*}{\textbf{FPS}} & \textbf{CPU} & \textbf{Estimated} & \textbf{Temp.} & \textbf{Discharge}\\
			& & \textbf{time (ms)} & \textbf{Power use (\%)} & \textbf{increase (ºC)} & \textbf{(mAh)}\\
			\hline
			BAD-256      & \second{29.970}& \first{73.4}  & \first{0.00026}  & \first{0.00030} & \first{0.0093} \\
			BAD-512      & \first{29.977} & \third{89.4}  & \third{0.00031}  & \second{0.00036} & \second{0.0104} \\
			BEBLID-512   & \third{29.91}  & 90.4          & \third{0.00031}  & \third{0.00037} & \third{0.0108} \\
			ORB          & 29.177         & \second{73.8} & \second{0.00028} & 0.00045 & 0.0118 \\
			\hline
			\hline
			SIFT        & 5.548  & 1062.1& 0.00337 & 0.00370 & 0.0798 \\
			HashSIFT-256    & \first{18.124} & \first{291.1} & \first{0.00097} & \first{0.00084} & \first{0.0217} \\
			HashSIFT-512    & \second{15.630} & \second{353.3} & \second{0.00119} & \second{0.00105} & \second{0.0262} \\
			BinBoost    & \third{9.841}  & \third{565.3} & \third{0.00180} & \third{0.00146} & \third{0.0408} \\ 
			\hline
			\hline
			Tfeat-m*    & \first{1.569} & \first{3441.9} & \first{0.01147} & \first{0.00968} & \first{0.2663} \\
			CDbin-256b  & \third{0.319} & \third{20195} & \third{0.06708} & \third{0.05396} & \third{1.3614} \\
			Hardnet     & \second{0.459} & \second{13099} & \second{0.04335} & \second{0.03561} & \second{0.9245} \\
			\hline
		\end{tabular}
	}
	\label{tab:battery}
\end{table}

In Table~\ref{tab:pipeline-times} we evaluate how significant is the description step (see Table~\ref{tab:execution_times}) compared with other steps in a typical application pipeline. We show the mean execution time in 100 runs of the detection, matching and model fitting steps involved in a homography estimation problem using OpenCV's implementation of ORB detection,
BFMatcher with Mutual Nearest Neighbors (MNN) and findHomography with PROSAC~\cite{chum2005prosac}. The replacement of ORB for BAD-256 not only increases the accuracy but also would represent a total reduction of about 30\% in the overall time. 
\begin{table}
    \centering
	\caption{Execution time (ms) of steps other than description in a homography estimation problem with OpenCV.}
	\begin{tabular}{|c|c|c|}\hline
		\textbf{Step} & \textbf{Intel Core i7} & \textbf{Snapdragon 855} \\ \hline
		ORB Detection & 13.33 ($\pm$0.28) & 17.02 ($\pm$0.16) \\
		Hamming MNN Matching & 4.65 ($\pm$ 0.16) & 9.50 ($\pm$ 0.26)\\
		PROSAC~\cite{chum2005prosac} & 0.69 ($\pm$ 0.02) & 0.45 ($\pm$ 0.0007)\\\hline
	\end{tabular}
    \label{tab:pipeline-times}
\end{table}

\section{Conclusion}

This paper introduces descriptors that establish new operation points in the state-of-the-art accuracy vs. efficiency curve. They significantly outperform ORB in terms of accuracy and are amenable to different resource budgets. 
We have revisited well-known efficient image features such as gray-level pixel differences and histograms of gradient orientation. And,
by adopting new learning techniques developed for DL, we boost the performance of descriptors built with them.

In a practical situation, if computational efficiency and energy consumption are our top priorities, we should use BAD-256.
It is three orders of magnitude more efficient than the most accurate binary descriptor, CDBin. The price we pay for this efficiency in a SfM problem in a complex scene, like Madrid Metropolis, is 26\% fewer points reconstructed. 
Compared to ORB we would have 28\% more sparse points. 
If we require more accuracy but we are still concerned about energy, we should use HashSIFT-512. It is one order of magnitude slower than BAD, but it increases its number of sparse points by 36\%. %
Moreover, in a planar registration problem, the mAP of HashSIFT-512 is better than the binary DL alternative and on par with floating-point approaches, 
but orders of magnitude more efficient. 

The state-of-the-art DL binary descriptor, CDbin, spends 146$\times$ more energy than our most efficient descriptor, BAD-256. This proves that the use of present DL-based descriptors entails a much larger energy budget. Therefore, the construction of more accurate but still very efficient descriptors is quite relevant for many robotics applications running on resource limited devices.
In summary, if resources are important then the proposed descriptors are the sweet spot. 

\ifCLASSOPTIONcaptionsoff
\newpage
\fi

\bibliographystyle{IEEEtran}
\bibliography{IEEEabrv,root}

\end{document}